%% file: main.tex
\documentclass[11pt,a4paper]{article}
\usepackage[hyperref]{emnlp2018}
\usepackage{standalone}
\usepackage{mystuff-nlp}
\usepackage{times}
\usepackage{tikz}
\usepackage{url}
\usepackage{latexsym}
\usepackage{color}
\usepackage{float}
\usepackage{amsfonts,amssymb,amsmath,bm,bbm}
\usepackage{algorithm}
\usepackage{algpseudocode}
\usepackage{graphicx,subfigure}
\usepackage{booktabs}
\usepackage{multirow}

\usepackage{enumitem,array}
\usepackage{setspace}

\newcommand{\example}[1]{`#1'}

\aclfinalcopy 

\title{Convolutional Neural Networks with Recurrent Neural Filters}

\author{Yi Yang\\
	Bloomberg\\
	New York, NY 10022\\
	{\tt yyang464@bloomberg.net}
        }

\date{}

\begin{document}
\maketitle
\begin{abstract}
\input{abstract}

\end{abstract}

\input{intro}

\input{model}

\input{exp}

\input{analysis}

\input{related}

\input{con}

\input{ack}

\bibliographystyle{acl_natbib_nourl}
\bibliography{cite-strings,cites,cite-definitions}

\end{document}

%% file: abstract.tex

We introduce a class of convolutional neural networks (CNNs) that utilize recurrent neural networks (RNNs) as convolution filters.
A convolution filter is typically implemented as a linear affine transformation followed by a non-linear function, 
which fails to account for language compositionality. As a result, it limits the use of high-order filters that are often warranted for natural language processing tasks.
In this work, we model convolution filters with RNNs that naturally capture compositionality and long-term dependencies in language.
We show that simple CNN architectures equipped with recurrent neural filters (RNFs) 
achieve results that are on par with the best published ones on the Stanford Sentiment Treebank and two answer sentence selection datasets.\footnote{The code is available at \url{https://github.com/bloomberg/cnn-rnf}.}

%% file: intro.tex
\section{Introduction}
\label{sec:intro}

Convolutional neural networks (CNNs) have been shown to achieve state-of-the-art results on various natural language processing (NLP) tasks, such as sentence classification~\cite{kim2014convolutional}, question answering~\cite{dong2015question}, and machine translation~\cite{gehring2017convolutional}. In an NLP system, a convolution operation is typically a sliding window function that applies a convolution filter to every possible window of words in a sentence. Hence, the key components of CNNs are a set of convolution filters that compose low-level word features into higher-level representations.  

Convolution filters are usually realized as linear systems, as their outputs are affine transformations of the inputs followed by some non-linear activation functions. 
Prior work directly adopts a linear convolution filter to NLP problems by utilizing a concatenation of embeddings of a window of words as the input, which leverages word order information in a shallow additive way. As early CNN architectures have mainly drawn inspiration from models of the primate visual system, the necessity of coping with language compositionality is largely overlooked in these systems. Due to the linear nature of the convolution filters, they lack the ability to capture complex language phenomena, such as compositionality and long-term dependencies.  

To overcome this, we propose to employ recurrent neural networks (RNNs) as convolution filters of CNN systems for various NLP tasks. Our recurrent neural filters (RNFs) can naturally deal with language compositionality with a recurrent function that models word relations, and they are also able to implicitly model long-term dependencies. RNFs are typically applied to word sequences of moderate lengths, which alleviates some well-known drawbacks of RNNs, including their vulnerability to the gradient vanishing and exploding problems~\cite{pascanu2013difficulty}. As in conventional CNNs, the computation of the convolution operation with RNFs can be easily parallelized. As a result, RNF-based CNN models can be $3$-$8$x faster than their RNN counterparts. 

We present two RNF-based CNN architectures for sentence classification and answer sentence selection problems. Experimental results on the Stanford Sentiment Treebank and the QASent and WikiQA datasets demonstrate that RNFs significantly improve CNN performance over linear filters by $4$-$5\%$ accuracies and $3$-$6\%$ MAP scores respectively. Analysis results suggest that RNFs perform much better than linear filters in detecting longer key phrases, which provide stronger cues for categorizing the sentences.



%% file: model.tex
\section{Approach}
\label{sec:model}

The aim of a convolution filter is to produce a local feature for a window of words.  We describe a novel approach to learning filters using RNNs, which is especially suitable for NLP problems. We then present two CNN architectures equipped with RNFs for sentence classification and sentence matching tasks respectively.

\subsection{Recurrent neural filters}
 
We briefly review the linear convolution filter implementation by~\newcite{kim2014convolutional}, which has been widely adopted in later works.  
Consider an $m$-gram word sequence $[ \mathbf{x}_i, \cdots, \mathbf{x}_{i+m-1} ]$, where $\mathbf{x}_i \in \mathbb{R}^k$ is a $k$-dimensional word vector. A linear convolution filter is a function applied to the $m$-gram to produce a feature $c_{i, j}$:
\begin{equation}
\begin{aligned}
c_{i, j} =& f(\mathbf{w}_j^\top \mathbf{x}_{i:i+m-1} + b_j), \\
\mathbf{x}_{i:i+m-1} =&  \mathbf{x}_i \oplus \mathbf{x}_{i+1} \oplus \cdots \oplus \mathbf{x}_{i+m-1},
\end{aligned}
\end{equation}
where $\oplus$ is the concatenation operator, $b_j$ is a bias term, and $f$ is a non-linear activation function. We typically use multiple independent linear filters to yield a feature vector $\mathbf{c}_i$ for the word sequence.
%
Linear convolution filters make strong assumptions about language that could harm the performance of NLP systems. First, linear filters assume local compositionality and ignore long-term dependencies in language. Second, they use separate parameters for each value of the time index, which hinders parameter sharing for the same word type~\cite{goodfellow2016deep}.
The assumptions become more problematic if we increase the window size $m$.

We propose to address the limitations by employing RNNs to realize convolution filters, which we term \emph{recurrent neural filters} (RNFs). RNFs compose the words of the $m$-gram from left to right using the same recurrent unit:
\begin{equation}
\mathbf{h}_{t} = \operatorname{RNN} (\mathbf{h}_{t-1}, \mathbf{x}_{t}),
\end{equation}
where $\mathbf{h}_t$ is a hidden state vector that encoded information about previously processed words, and the function $\operatorname{RNN}$ is a recurrent unit such as Long Short-Term Memory (LSTM) unit~\cite{hochreiter1997long} or Gated Recurrent Unit (GRU)~\cite{cho2014learning}. We use the last hidden state $ \mathbf{h}_{i+m-1}$ as the RNF output feature vector $\mathbf{c}_i$. Features learned by RNFs are interdependent of each other, which permits the learning of complementary information about the word sequence. The left-to-right word composing procedure in RNFs preserves word order information and implicitly models long-term dependencies in language. RNFs can be treated as simple drop-in replacements of linear filters and potentially adopted in numerous CNN architectures.



\subsection{CNN architectures}

\paragraph{Sentence encoder} We use a CNN architecture with one convolution layer followed by one max pooling layer to encode a sentence. In particular, the RNFs are applied to every possible window of $m$ words in the sentence $\{ \mathbf{x}_{1:m}, \mathbf{x}_{2:m+1}, \cdots, \mathbf{x}_{n-m+1:n}  \}$ to generate feature maps $\mathbf{C} = [\mathbf{c}_1, \mathbf{c}_2, \cdots, \mathbf{c}_{n-m+1}]$. Then a max-over-time pooling operation~\cite{collobert2011natural} is used to produce a fixed size sentence representation: $\mathbf{v} = \max \{ \mathbf{C} \}$.


\paragraph{Sentence classification} We use a CNN architecture that is similar to the CNN-static model described in~\newcite{kim2014convolutional} for sentence classification. After obtaining the sentence representation $\mathbf{v}$, a fully connected softmax layer is used to map $\mathbf{v}$ to an output probability distribution. The network is optimized against a cross-entropy loss function. 

\paragraph{Sentence matching} We exploit a CNN architecture that is nearly identical to the CNN-Cnt model introduced by~\newcite{yang2015wikiqa}. Let $\mathbf{v}_1$ and $\mathbf{v}_2$ be the vector representations of the two sentences. A bilinear function is applied to $\mathbf{v}_1$ and $\mathbf{v}_2$ to produce a sentence matching score. The score is combined with two word matching count features and fed into a sigmoid layer. The output of the sigmoid layer is used by binary cross-entropy loss to optimize the model. 

%% file: exp.tex
\section{Experiments}
\label{sec:exp}

We evaluate RNFs on some of the most popular datasets for the sentence classification and sentence matching tasks. After describing the experimental setup, we compare RNFs against both linear filters and conventional RNN models, and report our findings in~\autoref{sec:exp:results}.

\subsection{Experimental settings}

\paragraph{Data} 
We use the Stanford Sentiment Treebank~\cite{socher2013recursive} in our sentence classification experiments. We report accuracy results for both binary classification and fine-grained classification settings. Two answer sentence selection datasets, QASent~\cite{wang2007jeopardy} and WikiQA~\cite{yang2015wikiqa}, are adopted in our sentence matching experiments. We use MAP and MRR to evaluate the performance of answer sentence selection models.

\paragraph{Competitive systems} 
We consider CNN variants with linear filters and RNFs. For RNFs, we adopt two implementations based on GRUs and LSTMs respectively. We also compare against the following RNN variants: GRU, LSTM, GRU with max pooling, and LSTM with max pooling.\footnote{Max pooling performed better than mean pooling in our preliminary experiments.} We use the publicly available 300-dimensional GloVe vectors~\cite{pennington2014glove} pre-trained with $840$ billion tokens to initialize the word embeddings for all the models. The word vectors are fixed during downstream training. Finally, we report best published results for each dataset.\footnote{We exclude results obtained from systems using external resources beyond word embeddings.} 

\paragraph{Parameter tuning}
For all the experiments, we tune hyperparameters on the development sets and report results obtained with the selected hyperparameters on the test sets.  After the preliminary search, we choose the number of hidden units (feature maps) from $\{ 200, 300, 400 \}$, and the filter window width from $\{ 2, 3, 4, 5 \}$ for linear filters and from  $\{ 5, 6, 7, 8 \}$ for RNFs. Linear filters tend to perform well with small window widths, while RNFs work better with larger window widths. We apply dropout to embedding layers, pooling layers, RNN input layers, and RNN recurrent layers. The dropout rates are selected from $\{ 0, 0.2, 0.4 \}$, where $0$ indicates that dropout is not used for the specific layer.  Optimization is performed by Adam~\cite{kingma2015adam} with early stopping. We conduct random search with a budget of $100$ runs to seek the best hyperparameter configuration for each system. 

\subsection{Results}
\label{sec:exp:results}

\input{results-tab}

The evaluation results on sentiment classification and answer sentence selection are shown in~\autoref{tab:sentiment} and~\autoref{tab:qa} respectively. RNFs significantly outperform linear filters on both tasks. In fact, we find that CNN-RNF variants significantly outperform CNN-linear-filter on nearly every hyperparameter configuration in our experiments. On the Stanford Sentiment Treebank, CNN-RNF-LSTM outperforms CNN-linear-filter by $5.4\%$ and $3.9\%$ accuracies on the fine-grained and binary classification settings respectively. We observe similar performance boosts by adopting RNFs for CNNs on the QASent and WikiQA test sets. CNN-RNF-GRU improves upon CNN-linear-filter by $3.7\%$ MRR score on QASent and CNN-RNF-LSTM performs better than CNN-linear-filter by $6.1\%$ MAP score on WikiQA. In particular, CNN-RNF-LSTM achieves $53.4\%$ and $90.0\%$ accuracies on the fine-grained and binary sentiment classification tasks respectively, which match the state-of-the-art results on the Stanford Sentiment Treebank. CNN-RNF-LSTM also obtains competitive results on answer sentence selection datasets, despite the simple model architecture compared to state-of-the-art systems.

Conventional RNN models clearly benefit from max pooling, especially on the task of answer sentence selection. Like RNF-based CNN models, max-pooled RNNs also consist of two essential layers. The recurrent layer learns a set of hidden states corresponding to different time steps, and the max pooling layer extracts the most salient information from the hidden states. However, a hidden state in RNNs encodes information about all the previous time steps, while RNFs focus on detecting local features from a window of words that can be more relevant to specific tasks. As a result, RNF-based CNN models perform consistently better than max-pooled RNN models. 

CNN-RNF models are much faster to train than their corresponding RNN counterparts, despite they have the same numbers of parameters, as RNFs are applied to word sequences of shorter lengths and the computation is parallelizable. The training of CNN-RNF-LSTM models takes $10$-$20$ mins on the Stanford Sentiment Treebank, which is $3$-$8$x faster than the training of LSTM models, on an NVIDIA Tesla P100 GPU accelerator.

%% file: results-tab.tex
\begin{table}[t!]
\centering
\begin{tabular}{l c c }
    \toprule
     System & Fine-grained & Binary \\ \midrule
     \multicolumn{3}{l}{\it CNN variants} \\[2pt]
    CNN-linear-filter & 48.0 & 86.1 \\
    CNN-RNF-GRU & 53.0 & \textbf{90.0} \\
    CNN-RNF-LSTM & \textbf{53.4} & \textbf{90.0} \\[2pt]
    \multicolumn{3}{l}{\it RNN variants} \\[2pt] 
    GRU & 50.5 & 88.7 \\
    LSTM & 50.3 & 89.3 \\
    GRU-maxpool & 51.7 & 89.7 \\
    LSTM-maxpool & 51.6 & 89.8 \\ \midrule
    \multicolumn{3}{l}{\it Best published results} \\[2pt]
    \cite{lei2017deriving} & \underline{53.2} & \underline{89.9} \\
    \bottomrule
\end{tabular}
\caption{Accuracy results on the Stanford Sentiment Treebank. The best results obtained from our implementations are in \textbf{bold}. The best published results are \underline{underlined}.}
\label{tab:sentiment}
\end{table}

\begin{table*}[t!]
\centering
\begin{tabular}{l c c c c}
    \toprule
    \multirow{2}{*}{System} & \multicolumn{2}{c}{QASent} & \multicolumn{2}{c}{WikiQA} \\
    \cmidrule{2-5}
     & MAP & MRR & MAP & MRR \\ \midrule
     \multicolumn{5}{l}{\it CNN variants} \\[2pt]
    CNN-linear-filter & 0.750 & 0.813 & 0.668 & 0.687 \\
    CNN-RNF-GRU & 0.773 & \textbf{0.850} & 0.726 & 0.738 \\
    CNN-RNF-LSTM & \textbf{0.780} & 0.841 & \textbf{0.729} & \textbf{0.747} \\[2pt]
    \multicolumn{5}{l}{\it RNN variants} \\[2pt] 
    GRU & 0.739 & 0.815 & 0.658 & 0.678 \\
    LSTM & 0.745 & 0.806 & 0.651 & 0.671 \\
    GRU-maxpool & 0.744 & 0.814 & 0.712 & 0.724 \\
    LSTM-maxpool & 0.762 & 0.825 & 0.701 & 0.711 \\ \midrule
    \multicolumn{3}{l}{\it Best published results} \\[2pt]
    \cite{rao2016noise} & \underline{0.801} & \underline{0.877} & 0.701 & 0.718 \\
    \cite{wang2017compare} & - & - & \underline{0.743} & \underline{0.755} \\
    \bottomrule
\end{tabular}
\caption{Evaluation results on answer sentence selection datasets. The best results obtained from our implementations are in \textbf{bold}. The best published results are \underline{underlined}.}
\label{tab:qa}
\end{table*}

%% file: analysis.tex
\section{Analysis}
\label{sec:analysis}

We now investigate why RNFs are more effective than linear convolution filters on the binary sentiment classification task. We perform quantitative analysis on the development set of the Stanford Sentiment Treebank (SST), in which sentiment labels for some phrases are also available.

\begin{figure}[t!]
\centering
\includegraphics[scale=.34]{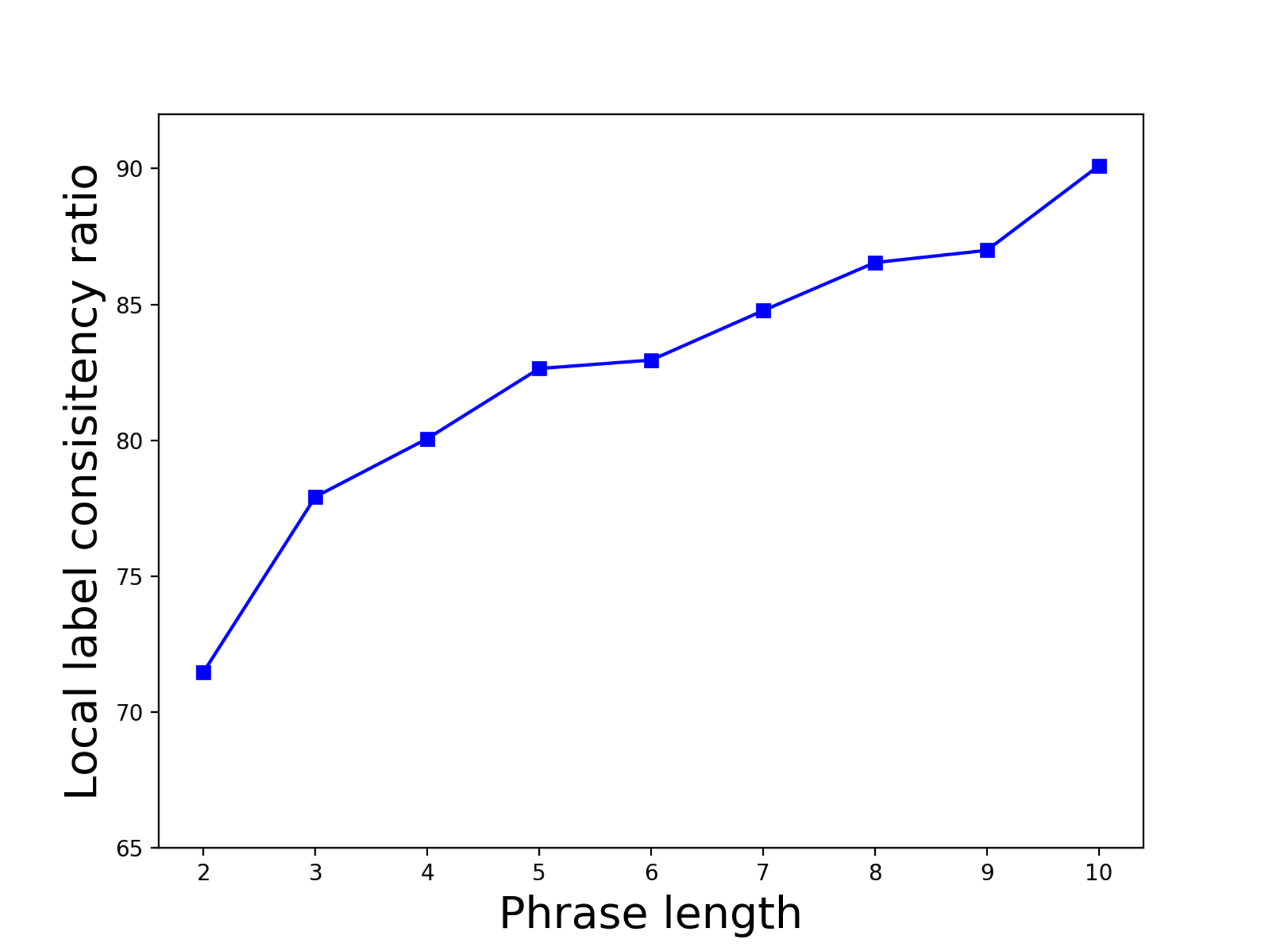}
\caption{Local label consistency ratio results computed on SST dev set.}
\label{fig:llc}
\end{figure}

\paragraph{Local label consistency (LLC) ratio}
We first inspect whether longer phrases have a higher tendency to express the same sentiment as the entire sentence. We define the \emph{local label consistency (LLC) ratio} as the ratio of $m$-grams that share the same sentiment labels as the original sentences. The LLC ratios with respect to different phrase lengths are shown in~\autoref{fig:llc}. Longer phrases are more likely to convey the same sentiments as the original sentences. Therefore, the ability to model long phrases is crucial to convolution filters.

\begin{figure}[t!]
\centering
\includegraphics[scale=.4]{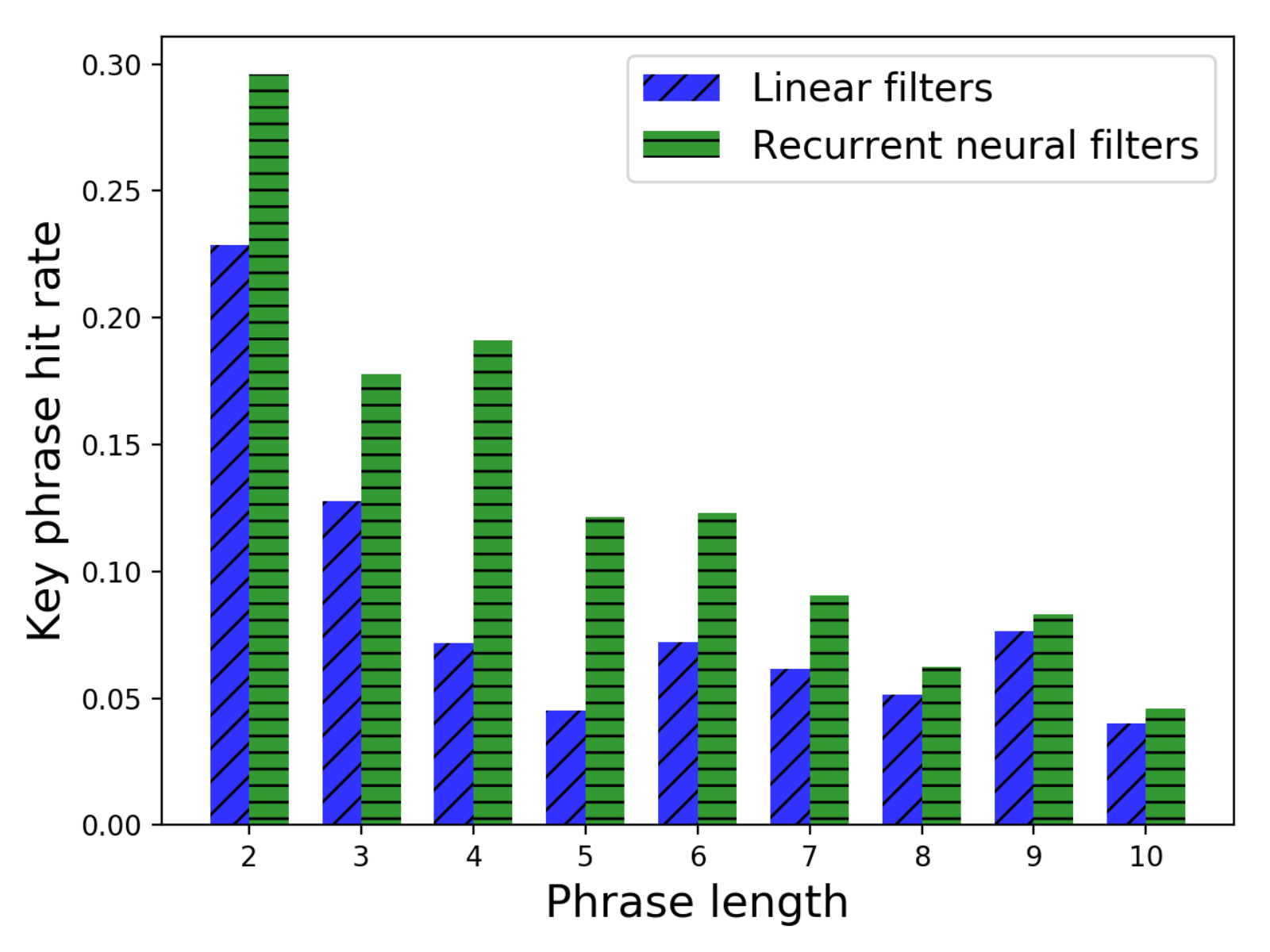}
\caption{Key phrase hit rate results computed on SST dev set. The RNFs are implemented with LSTMs.}
\label{fig:hit}
\end{figure}

\paragraph{Key phrase hit rate}
We examine linear filters and RNFs on the ability to detect a key phrase in a sentence. Specifically, we define the key phrases for a sentence to be the set of phrases that are labeled with the same sentiment as the original sentence. The \emph{key phrase hit rate} is then defined as the ratio of filter-detected $m$-grams that fall into the corresponding key phrase sets. The filter-detected $m$-gram of a sentence is the one whose convolution feature vector has the shortest Euclidean distance to the max-pooled vector. 
The hit rate results computed on SST dev set are presented in~\autoref{fig:hit}. As shown, RNFs consistently perform better than linear filters on identifying key phrases across different phrase lengths, especially on phrases of moderate lengths ($4$-$7$).  The results suggest that RNFs are superior to linear filters, as they can better model longer phrases whose labels are more consistent with the sentences.\footnote{Phrases of very long lengths ($8$-$10$) are rarely annotated in SST dev data, which could explain why linear filters and RNFs achieve similar hit rates, as small data sample often leads to high variance.}

%% file: related.tex
\section{Related work}
\label{sec:related}

Linear convolution filters are dominated in CNN-based systems for both computer vision and natural language processing tasks. One exception is the work of~\newcite{zoumpourlis2017non}, which proposes a convolution filter that is a function with quadratic forms through the Volterra kernels. However, this non-linear convolution filter is developed in the context of a computational model of the visual cortex, which is not suitable for NLP problems. In contrast, RNFs are specifically derived for NLP tasks, in which a different form of non-linearity, language compositionality, often plays a critical role.

Several works have employed neural network architectures that contain both CNN and RNN components to tackle NLP problems. \newcite{tan2015lstm} present a deep neural network for answer sentence selection, in which a convolution layer is applied to the output of a BiLSTM layer for extracting sentence representations. \newcite{ma2016end} propose to compose character representations of a word using a CNN, whose output is then fed into a BiLSTM for sequence tagging. 
\newcite{kalchbrenner2013recurrent} introduce a neural architecture that uses a sentence model based on CNNs and a discourse model based on RNNs. Their system achieves state-of-the-art results on the task of dialogue act classification. 
Instead of treating an RNN and a CNN as isolated components, our work directly marries RNNs with the convolution operation, which illustrates a new direction in mixing RNNs with CNNs.

%% file: con.tex
\section{Conclusion and future work}
\label{sec:con}

We present RNFs, a new class of convolution filters based on recurrent neural networks. RNFs sequentially apply the same recurrent unit to words of a phrase, which naturally capture language compositionality and implicitly model long-term dependencies. Experiments on sentiment classification and answer sentence selection tasks demonstrate that RNFs give a significant boost in performance compared to linear convolution filters. RNF-based CNNs also outperform a variety of RNN-based models, as they focus on extracting local information that could be more relevant to particular problems. The quantitive analysis reveals that RNFs can handle long phrases much better than linear filters, which explains their superiority over the linear counterparts. In the future, we would like to investigate the effectiveness of RNFs on a wider range of NLP tasks, such as natural language inference and machine translation.

%% file: ack.tex
\section{Acknowledgments}

We thank Chunyang Xiao for helping us to run the analysis experiments. We thank Kazi Shefaet Rahman, Ozan Irsoy, Chen-Tse Tsai, and Lingjia Deng for their valuable comments on
earlier versions of this paper. We also thank the EMNLP reviewers for their helpful feedback. 

%% file: main.bbl
\begin{thebibliography}{20}
\expandafter\ifx\csname natexlab\endcsname\relax\def\natexlab#1{#1}\fi

\bibitem[{Cho et~al.(2014)Cho, Van~Merri{\"e}nboer, Gulcehre, Bahdanau,
  Bougares, Schwenk, and Bengio}]{cho2014learning}
Kyunghyun Cho, Bart Van~Merri{\"e}nboer, Caglar Gulcehre, Dzmitry Bahdanau,
  Fethi Bougares, Holger Schwenk, and Yoshua Bengio. 2014.
\newblock Learning phrase representations using rnn encoder-decoder for
  statistical machine translation.
\newblock In \emph{{Proceedings of Empirical Methods for Natural Language
  Processing (EMNLP)}}.

\bibitem[{Collobert et~al.(2011)Collobert, Weston, Bottou, Karlen, Kavukcuoglu,
  and Kuksa}]{collobert2011natural}
Ronan Collobert, Jason Weston, L{\'e}on Bottou, Michael Karlen, Koray
  Kavukcuoglu, and Pavel Kuksa. 2011.
\newblock Natural language processing (almost) from scratch.
\newblock \emph{Journal of Machine Learning Research}, 12(Aug):2493--2537.

\bibitem[{Dong et~al.(2015)Dong, Wei, Zhou, and Xu}]{dong2015question}
Li~Dong, Furu Wei, Ming Zhou, and Ke~Xu. 2015.
\newblock Question answering over freebase with multi-column convolutional
  neural networks.
\newblock In \emph{{Proceedings of the Association for Computational
  Linguistics (ACL)}}.

\bibitem[{Gehring et~al.(2017)Gehring, Auli, Grangier, Yarats, and
  Dauphin}]{gehring2017convolutional}
Jonas Gehring, Michael Auli, David Grangier, Denis Yarats, and Yann~N Dauphin.
  2017.
\newblock Convolutional sequence to sequence learning.
\newblock In \emph{{Proceedings of the International Conference on Machine
  Learning (ICML)}}.

\bibitem[{Goodfellow et~al.(2016)Goodfellow, Bengio, Courville, and
  Bengio}]{goodfellow2016deep}
Ian Goodfellow, Yoshua Bengio, Aaron Courville, and Yoshua Bengio. 2016.
\newblock \emph{Deep learning}, volume~1.
\newblock MIT press Cambridge.

\bibitem[{Hochreiter and Schmidhuber(1997)}]{hochreiter1997long}
Sepp Hochreiter and J{\"u}rgen Schmidhuber. 1997.
\newblock Long short-term memory.
\newblock \emph{Neural computation}, 9(8).

\bibitem[{Kalchbrenner and Blunsom(2013)}]{kalchbrenner2013recurrent}
Nal Kalchbrenner and Phil Blunsom. 2013.
\newblock Recurrent convolutional neural networks for discourse
  compositionality.
\newblock In \emph{Proceedings of the Workshop on Continuous Vector Space
  Models and their Compositionality}.

\bibitem[{Kim(2014)}]{kim2014convolutional}
Yoon Kim. 2014.
\newblock Convolutional neural networks for sentence classification.
\newblock In \emph{{Proceedings of Empirical Methods for Natural Language
  Processing (EMNLP)}}.

\bibitem[{Kingma and Ba(2015)}]{kingma2015adam}
Diederik~P Kingma and Jimmy Ba. 2015.
\newblock Adam: A method for stochastic optimization.
\newblock In \emph{{Proceedings of the International Conference on Learning
  Representations (ICLR)}}.

\bibitem[{Lei et~al.(2017)Lei, Jin, Barzilay, and Jaakkola}]{lei2017deriving}
Tao Lei, Wengong Jin, Regina Barzilay, and Tommi Jaakkola. 2017.
\newblock Deriving neural architectures from sequence and graph kernels.
\newblock In \emph{{Proceedings of the International Conference on Machine
  Learning (ICML)}}.

\bibitem[{Ma and Hovy(2016)}]{ma2016end}
Xuezhe Ma and Eduard Hovy. 2016.
\newblock End-to-end sequence labeling via bi-directional lstm-cnns-crf.
\newblock In \emph{{Proceedings of the Association for Computational
  Linguistics (ACL)}}.

\bibitem[{Pascanu et~al.(2013)Pascanu, Mikolov, and
  Bengio}]{pascanu2013difficulty}
Razvan Pascanu, Tomas Mikolov, and Yoshua Bengio. 2013.
\newblock On the difficulty of training recurrent neural networks.
\newblock In \emph{{Proceedings of the International Conference on Machine
  Learning (ICML)}}.

\bibitem[{Pennington et~al.(2014)Pennington, Socher, and
  Manning}]{pennington2014glove}
Jeffrey Pennington, Richard Socher, and Christopher Manning. 2014.
\newblock Glove: Global vectors for word representation.
\newblock In \emph{{Proceedings of Empirical Methods for Natural Language
  Processing (EMNLP)}}.

\bibitem[{Rao et~al.(2016)Rao, He, and Lin}]{rao2016noise}
Jinfeng Rao, Hua He, and Jimmy Lin. 2016.
\newblock Noise-contrastive estimation for answer selection with deep neural
  networks.
\newblock In \emph{Proceedings of the ACM International on Conference on
  Information and Knowledge Management (CIKM)}.

\bibitem[{Socher et~al.(2013)Socher, Perelygin, Wu, Chuang, Manning, Ng, and
  Potts}]{socher2013recursive}
Richard Socher, Alex Perelygin, Jean Wu, Jason Chuang, Christopher~D Manning,
  Andrew Ng, and Christopher Potts. 2013.
\newblock Recursive deep models for semantic compositionality over a sentiment
  treebank.
\newblock In \emph{{Proceedings of Empirical Methods for Natural Language
  Processing (EMNLP)}}.

\bibitem[{Tan et~al.(2016)Tan, Santos, Xiang, and Zhou}]{tan2015lstm}
Ming Tan, Cicero~dos Santos, Bing Xiang, and Bowen Zhou. 2016.
\newblock Lstm-based deep learning models for non-factoid answer selection.
\newblock In \emph{{Proceedings of the International Conference on Learning
  Representations (ICLR)}}.

\bibitem[{Wang et~al.(2007)Wang, Smith, and Mitamura}]{wang2007jeopardy}
Mengqiu Wang, Noah~A Smith, and Teruko Mitamura. 2007.
\newblock What is the jeopardy model? a quasi-synchronous grammar for qa.
\newblock In \emph{{Proceedings of Empirical Methods for Natural Language
  Processing (EMNLP)}}.

\bibitem[{Wang and Jiang(2017)}]{wang2017compare}
Shuohang Wang and Jing Jiang. 2017.
\newblock A compare-aggregate model for matching text sequences.
\newblock In \emph{{Proceedings of the International Conference on Learning
  Representations (ICLR)}}.

\bibitem[{Yang et~al.(2015)Yang, Yih, and Meek}]{yang2015wikiqa}
Yi~Yang, Wen-tau Yih, and Christopher Meek. 2015.
\newblock Wikiqa: A challenge dataset for open-domain question answering.
\newblock In \emph{{Proceedings of Empirical Methods for Natural Language
  Processing (EMNLP)}}.

\bibitem[{Zoumpourlis et~al.(2017)Zoumpourlis, Doumanoglou, Vretos, and
  Daras}]{zoumpourlis2017non}
Georgios Zoumpourlis, Alexandros Doumanoglou, Nicholas Vretos, and Petros
  Daras. 2017.
\newblock Non-linear convolution filters for cnn-based learning.
\newblock In \emph{Proceedings of the International Conference on Computer
  Vision (ICCV)}.

\end{thebibliography}
